\documentclass[final,3p,times]{sarticle}

\usepackage{amssymb}
\usepackage{amsmath,amsfonts}
\usepackage{bbm}
\usepackage[linesnumbered,ruled,vlined]{algorithm2e}
\usepackage{array}
\usepackage[caption=false,font=normalsize,labelfont=sf,textfont=sf]{subfig}
\usepackage{textcomp}
\usepackage{stfloats}
\usepackage{url}
\usepackage{verbatim}
\usepackage{graphicx}
\usepackage{booktabs}
\usepackage{rotating}
\usepackage{makecell}
\usepackage{color}
\usepackage{longtable}
\usepackage{tabularx}
\usepackage{amssymb, mathrsfs}
\usepackage{picins}
\usepackage{multirow}
\usepackage{setspace}
\usepackage{geometry}
\usepackage{hyperref}  
\hypersetup{hidelinks}
\biboptions{sort&compress}
\begin{document}
\begin{sloppypar}

\begin{frontmatter}


\title{Pseudolabel guided pixels contrast for domain adaptive semantic segmentation}
\author{Jianzi Xiang, Cailu Wan, and Zhu Cao$^{*}$}
\address{The Key Laboratory of Smart Manufacturing in Energy Chemical Process, Ministry of Education, East China University of Science and Technology, Shanghai 200237, China

*Corresponding author email: caozhu@ecust.edu.cn}


\begin{abstract}
	Semantic segmentation is essential for comprehending images, but the process necessitates a substantial amount of detailed annotations at the pixel level. Acquiring such annotations can be costly in the real-world. Unsupervised domain adaptation (UDA) for semantic segmentation is a technique that uses virtual data with labels to train a model and adapts it to real data without labels. Some recent works use contrastive learning, which is a powerful method for self-supervised learning, to help with this technique. However, these works do not take into account the diversity of features within each class when using contrastive learning, which leads to errors in class prediction. We analyze the limitations of these works and propose a novel framework called \emph{P}seudo-label \emph{G}uided \emph{P}ixel \emph{C}ontrast (PGPC), which overcomes the disadvantages of previous methods. We also investigate how to use more information from target images without adding noise from pseudo-labels. We test our method on two standard UDA benchmarks and show that it outperforms existing methods. Specifically, we achieve relative improvements of $5.1\%$ mIoU and $4.6\%$ mIoU on the Grand Theft Auto V (GTA5) to Cityscapes and SYNTHIA to Cityscapes tasks based on DAFormer, respectively. Furthermore, our approach can enhance the performance of other UDA approaches without increasing model complexity. Code is available at https://github.com/embar111/pgpc
\end{abstract}

\begin{keyword}
Semantic Segmentation \sep Unsupervised Domain Adaptation \sep Contrastive Learning


\end{keyword}

\end{frontmatter}


\section{Introduction} \label{introduction}
Semantic segmentation, assigning semantic labels to pixels, is crucial for applications like medical analysis \cite{unet_biomedical}, robot navigation \cite{zurbrugg2022embodied}, and autopilot \cite{autonomous_driving}. In the preceding years, numerous methods founded on Convolutional Neural Networks (CNN) \cite{deeplabv2,deeplabv3,li2020learning, MLFNet, sun2024extraction} and Transformers \cite{lu2022deformable, setr, segformer, miao2024efficient} have been proposed and obtained surprising effects. However, these methods rely on a mass of expensive and labor-intensive pixel-level labels thus hindering their practical applicability. One alternative to solve this problem is training semantic segmentation networks using virtual datasets \cite{GTA, synthia} that can automatically obtain pixel-level labels. Therefore, the problem of eliminating the domain gap between virtual (source domain) and real (target domain) images, known as UDA on semantic segmentation, has garnered significant interest among researchers.

Recent works for UDA attempt to learn the common domain-invariant knowledge through two methods: adversarial learning \cite{pixelIntraDA} and self-training \cite{DAP, uncertainty_aware, PDA, zhang2020knowledge, ren2022multi, lin2022prototype}. Adversarial learning aligns global distributions between the source and target domain by deceiving a domain discriminator, but it fails to ensure the separability of features across different categories in the target domain \cite{yang2020phase}. Self-training methods, which adopted confidence estimation \cite{confidence_estimation}, consistency regularization \cite{consistency_regularization}, or entropy minimization \cite{entropy_minimization} to improve the segmentation performance, utilize pseudo-labels to enable a class-ware alignment. Despite the good performance of these methods, they mostly do not accurately bridge domain discrepancies, resulting in indistinct target representations \cite{cluda}.
To address the problem described above, motivated by the success of contrastive learning \cite{chopra2005learning} whose objective is to separate dissimilar (negative) data pairs and align similar (positive) data pairs in the self-supervised learning, some recent works \cite{sepico, cluda, prototypical_contrast, huang2022category, lee2022bi} apply contrastive learning to the UDA on semantic segmentation and obtain good results. 

\begin{figure}[th]
	\centering
	\includegraphics[scale=1]{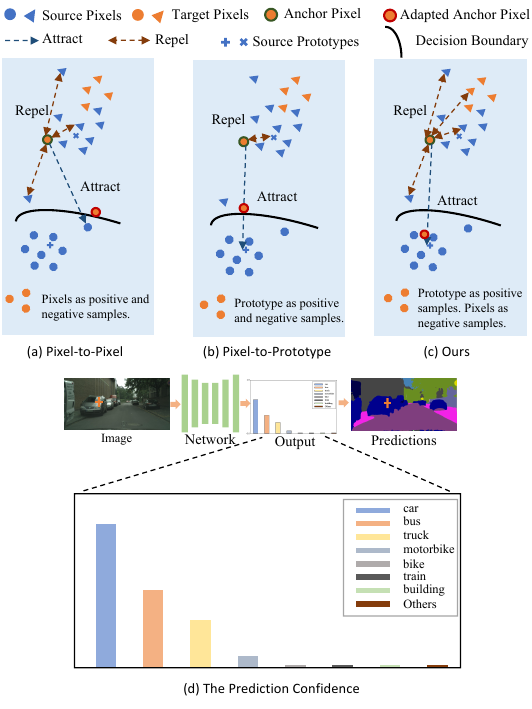}
	\doublespacing\caption{The schematic diagram of our motivation. (a) and (b) are the pixel-to-pixel and pixel-to-prototype contrastive learning methods, respectively. (c) is our contrastive learning method. (d) is the prediction confidence of the pixel marked with an orange cross. The adapted anchor pixel means the anchor pixel is trained by contrastive learning. The adapted anchor pixel is misplaced on the incorrect side of the decision boundary in (a) and (b), while (c) correctly positions it on the right side.}
	\label{Introduction_insight}
\end{figure}

Currently, there exist opportunities for enhancement in two aspects of these contrastive methods: the modality of contrastive learning and the utilization of pseudo-labels. On the one hand, contrastive learning in UDA for semantic segmentation primarily employs two methodological approaches: pixel-to-pixel and  pixel-to-prototype. The pixel-to-pixel methods \cite{cluda, huang2022category} use contrastive learning to narrow the gap within pixel-wise feature representations for pixels sharing the same class, while simultaneously widening the separation for those belonging to distinct classes, irrespective of the domain. However, these methods do not guarantee that the pixel representations are close to the feature center of their category, because the pixel representations are scattered in different spatial locations, as shown in Fig. \ref{Introduction_insight}(a). The pixel-to-prototype methods \cite{sepico, lee2022bi} use contrastive learning to reduce the discrepancies between pixel representations and their respective category prototypes, which are the feature centers of each category. These methods are effective in tackling the challenges faced by pixel-to-pixel methods, but they do not separate the pixel representations of different categories well enough across domains. This is because each category has a large diversity of pixel representations, and the feature boundaries of each category may be far from the category prototypes.  Even when the pixel features of one class have been distanced from the prototypes of other classes, they might still reside within the feature spaces of those other classes, as depicted in Fig. \ref{Introduction_insight}(b). So the above two types of contrastive learning methods may still fail to learn distinguishable representations. 

Conversely, the process of semantic segmentation necessitates pixel-wise guidance for the execution of contrastive learning. However, target images do not have ground-truth labels, so previous methods use pseudo-labels generated from predictions to perform contrastive learning at the pixel level. Some methods \cite{sepico} use all predictions as pseudo-labels without filtering out noisy pseudo-labels, which can degrade the model performance. To avoid confirmation bias \cite{arazo2020pseudo} and reduce the effect of incorrect pseudo-labels, some methods \cite{prototypical_contrast} use a class-wise adaptive thresholding method to select only confident predictions as pseudo-labels and ignore uncertain ones. But insufficient use of pixels in the target image can also lead to degradation of the final performance. Our method selects reliable predictions as pseudo-labels, thereby minimizing the influence of noise. With each training iteration, the model updates its predictions, yielding even more credible pseudo-labels. Through this iterative process, spanning the entire training duration, the model harnesses an increasingly vast array of pixels for learning, ultimately enhancing its overall performance. As a prominent approach within unsupervised learning, our pseudo-label methodology prioritizes utilizing the utmost quantity of reliable pseudo-label samples as training data, thereby offering a valuable benchmark for subsequent endeavors in this domain.

Through these analyses, we introduce an innovative framework that integrates both the pixel-to-pixel and pixel-to-prototype methods first. 
Specifically, our method makes use of class prototypes as positive instances to reduce their distance from the prototype of the same category 
in the source domain for each target pixel representation of a certain category, and pixels to increase its distance to some random source 
pixel representations of other categories. By integrating the pixel-to-pixel method with the pixel-to-prototype approach, our model gains the capability 
to not only discern pixel representations and the distinctions between pixels belonging to diverse categories but also capture the similarities between pixel 
representations and the prototypes of the same category.
Then, we propose a method that leverages all target pixels in contrastive learning for UDA. 
The method is derived from the assumption that the model can differentiate between dissimilar classes and only confuse a few classes that are similar. 
In particular, as indicated in Fig. \ref{Introduction_insight}(d), a pixel may have similar probabilities for class \(car\), \(bus\), and \(truck\). 
But the model confidently rejects class \(person\), \(vegetable\), and other classes. So we can use this pixel as a negative sample for class \(person\) or 
\(vegetable\). We combine this method with our proposed framework to achieve better results. 
For example, as shown in Fig.~\ref{Introduction_insight}(c), our method reduces the distance between the anchor pixel and positive samples (prototype), while increasing the distance to negative samples (pixels), so that it is classified as the correct category. This approach enables the simultaneous consideration of both the wide distribution of pixels in the feature space and the excessive distance between prototypes and decision boundaries.
Below is a detailed account of the primary contributions made by our study:
\begin{itemize}
	\item {We propose a new contrastive learning framework for UDA on semantic segmentation by integrating the pixel-to-pixel and pixel-to-prototype methods.}
	\item {We introduce a novel approach, which can utilize all target pixels effectively in our proposed framework.} 
	\item {We perform comprehensive evaluations on renowned UDA benchmarks, specifically the GTA5 to Cityscapes and Synthia to Cityscapes transitions. Our approach has demonstrated significant improvements across various advanced UDA frameworks. For example, we attain relative improvements of 5.1\% mIoU and 4.6\% mIoU on the two UDA benchmarks based on DAFormer, respectively.}
\end{itemize}

\section{Related Work}
\subsection{Semantic Segmentation} 
The field of semantic segmentation has experienced remarkable progress since Long \textit{et~al.} \cite{FCN} utilized CNN for this task. A common challenge for semantic segmentation is to capture large-scale contextual information, which requires large receptive fields. In response to this issue, numerous strategies have been put forward, such as context aggregation \cite{PSP, MultiScale}, attention modules \cite{OCR, dual_attention, sparse_attention}, and atrous convolution \cite{deeplabv2, deeplabv3}. Complementing CNN-based techniques, the Transformer-based methods have received considerable attention from researchers in recent times. These Transformer-based \cite{transformer} models have demonstrated exceptional efficacy in the realm of natural language processing, and have inspired recent efforts to apply them to visual tasks \cite{ViT, setr, segformer, swin}. One of the pioneering methods in this direction is SETR \cite{setr} which adopts a sequence-to-sequence approach and uses ViT \cite{ViT}, a framework purely based on Transformers, as its core for extracting features from image tokens. ViT divides an image into token sequences and processes them with a series of Transformer encoders, demonstrating the effectiveness of Transformers for visual recognition. Moreover, Swin Transformer \cite{swin} introduces a hierarchical Transformer that can capture multiscale visual information and adapt to the differences between language and vision domains. SegFormer \cite{segformer} designs an innovative encoder with a hierarchical Transformer architecture that produces features across various scales and a lightweight decoder that combines them into a powerful representation. These Transformer-based methods consistently achieve remarkable results on semantic segmentation.
\subsection{Unsupervised Domain Adaptation} 
Unsupervised domain adaptation is a technique that uses virtual data with labels to train a model and adapts it to real data without labels. Semantic segmentation requires a substantial number of pixel-level annotations, but acquiring these annotations in the real world can be quite costly, so UDA is suitable for semantic segmentation that require a large amount of pixel-level annotations. The landscape of UDA techniques for semantic segmentation typically encompasses two primary strategies: methods grounded in adversarial training and those based on self-training approaches. Adversarial training approaches regard the segmentation modules as generators that try to produce domain-invariant features \cite{FcnsInTheWild} or predictions \cite{pixelIntraDA} for images from different domains, while discriminators are utilized to identify the domain of the features or predictions. Alternatively, UDA can be achieved through the process of imparting the visual aesthetics of source domain to the target domain, leveraging the principles of adversarial training as referenced in \cite{cycada, kim2020learning}. Self-training techniques are designed to generate pseudo-labels for target images, leveraging a network trained on labeled source images. However, domain discrepancy inevitably introduces noise into these pseudo-labels. In response to this challenge, numerous strategies have been put forward, such as confidence thresholding \cite{IAST, CRST}, consistency regularization \cite{fixmatch, consistency_regularization}, domain-mixup \cite{DSP, hoyer2021improving}, and pseudo-labels rectification \cite{uncertainty_aware,IR2F-RMM}. Self-training methods have emerged as the dominant methods in UDA due to their stability and superior performance. DAFormer \cite{daformer} is the first method to integrate the Transformer architecture and self-training for UDA. To ensure a stable training process and prevent overfitting to the source domain, DAFormer employs three simple yet effective training strategies. Building on this work, HRDA \cite{hrda} leverages high-quality and low-quality images to enhance the segmentation results through a scale attention module, leading to top-tier performance levels.
\subsection{Contrastive Learning} 
The contrastive learning approach \cite{chopra2005learning} aims to minimize the differences in features among similar instances, while simultaneously maximizing the distinctions between dissimilar ones. It is a key component of metric learning and has been widely used in self-supervised learning \cite{chen2020big, chen2021empirical, grill2020bootstrap}. Some recent works \cite{region_aware_contrastive_learning, zhong2021pixel, lai2021semi, u2pl, wang2021exploring} have applied contrastive learning for semantic segmentation with pixel-level labels in either a fully supervised or a semi-supervised setting. These methods have achieved favorable outcomes in both fully supervised and semi-supervised domains. Using these techniques, one can extract more robust and discriminative features. However, since these methods do not account for cross-domain adaptation concerns, they cannot be readily applied in unsupervised domains. Hence, several investigations have explored integrating contrastive learning in unsupervised semantic segmentation and have achieved notable performance. Currently, there are three types of methods: pixel-to-pixel \cite{huang2022category, cluda}, pixel-to-prototype \cite{sepico, lee2022bi}, and prototype-to-prototype \cite{prototypical_contrast}. These methods are capable of reducing domain discrepancies while simultaneously enhancing the discriminability among different categories.

\section{Method}
Below, Section \ref{sec_III_A} introduces the mathematical framework that defines the issue at hand and an overview of PGPC. Section \ref{sec_III_B} describes the strategy for selecting reliable target image pixels. Section \ref{sec_III_C} explains the specific contrastive learning algorithm of PGPC.

\subsection{Overview} \label{sec_III_A}
Considering the problem of UDA for semantic segmentation involves a collection of source images $\mathcal{X}_s = {\{x_s^{i}\}}_{i=1}^{N_s}$ from the virtual-world with $\mathit{C}$-class segmentation labels $\mathcal{Y}_s = {\{y_s^{i}\}}_{i=1}^{N_s}$, and a collection of target images $\mathcal{X}_t={\{x_t^{i}\}}_{i=1}^{N_t}$ from the real-world without any labels. We aim to develop a neural network model capable of utilizing the source data to boost the precision of segmentation when applied to target data.

\begin{figure}[thbp]
	\centering
	\includegraphics[scale=1]{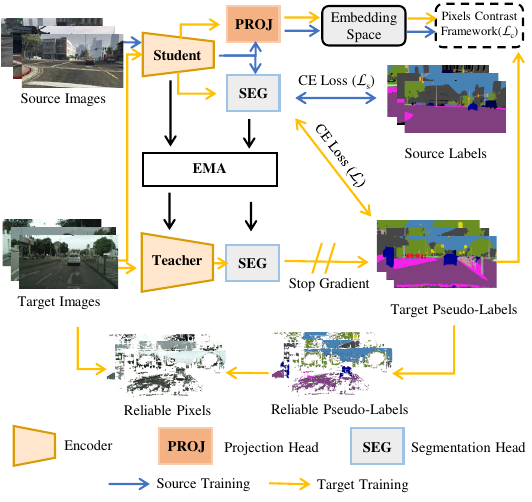}
	\doublespacing\caption{An overview of PGPC. The student model, guided by the fundamental segmentation loss \(\mathcal{L}_s\), produces predictions on the source data. Concurrently, the teacher model provides an estimation of the pseudo-labels for the target data. Subsequently, the student model issues predictions on the target data, under the oversight of the weighted segmentation loss \(\mathcal{L}_t\). In addition to the segmentation losses, the contrast loss \(\mathcal{L}_c\) is introduced to encourage the alignment of pixel features in the embedding space.}
	\label{network} 
\end{figure}

Figure \ref{network} illustrates our method, which employs a self-training paradigm featuring a pair of models with matching designs, known as the teacher and student models. A set \(B\) of labeled source images, $\mathcal{B}_s$, and an equivalent set of unlabeled target images, $\mathcal{B}_t$, are selected at each step \(t\) in the training process. The student model with parameters $\Theta$
undergoes preliminary training on the source images by focusing on decreasing the categorical cross-entropy $(CE)$ loss, which is a fully supervised objective:

\begin{equation}
\begin{aligned}
\mathcal{L}_{s} = - \frac{1}{B} \sum\limits_{(x_s^i, y_s^i) \in \mathcal{B}_s} \sum\limits_{j=1}^{H \times W} \sum\limits_{c=1}^{C} (y_s^i)^{(j,c)}\log{\Theta (x_s^i)}^{(j,c)},
\label{source supervised loss}
\end{aligned}
\end{equation}
where \(H\) and \(W\) denote the image vertical extent and horizontal span. To enhance the knowledge transfer across domains, self-training methods employ the teacher model $\Theta'$ that produces pseudo-labels of the images in the target domain:
\begin{equation}
\begin{aligned}
(p_t^i)^{(j,c')} = \mathop{\mathrm{argmax}}\limits_{c'} \Theta' (x_t^i)^{(j,c')}, j \in \{1,2, \dots, H \times W\}.
\label{pseudo label}
\end{aligned}
\end{equation}
Note that the teacher model $\Theta'$ undergoes updates by taking into account the exponential moving average \((EMA)\) of weights associated with the student model.
 Here, we use the ratio of pixels exceeding a threshold $\kappa$ (an adaptive value that can be learned through training) of the maximum softmax probability \cite{daformer}.
\begin{equation}
	\begin{aligned}
		q_T^{(i)} = \frac{\sum\limits_{j=1}^{H \times W}[\mathop{\mathrm{max}}\limits_{c'} \Theta'(x_T^{(i)})^{(j,c')}>\kappa]}{H \times W}
	\end{aligned}
\end{equation}

Subsequently, the student model $\Theta$ is fine-tuned on the target domain using the CE loss function with the guidance of pseudo-labels:

\begin{equation}
\begin{aligned}
\mathcal{L}_{t} = - \sum\limits_{j=1}^{H \times W} \sum\limits_{c=1}^{C} q_T^{(i)} (p_t^i)^{(j,c)}\log{\Theta (x_t^i)}^{(j,c)}.
\label{target self-training loss}
\end{aligned}
\end{equation}

Within the framework of self-training, we employ the contrastive loss to improve the feature discriminability among diverse classes. $\mathcal{L}_c$ is the pixel-wise InfoNCE \cite{InfoNCE} computed as:
\begin{equation}
\begin{aligned}
\mathcal{L}_c = & - \frac{1}{C \times M} \sum\limits_{c=0}^{C-1}\sum\limits_{m=1}^M \\
& \log \left[ \frac{e^{\langle z_{cm}, z_{c}^+ \rangle / \tau}}{e^{\langle z_{cm}, z_{c}^+ \rangle / \tau} + \sum\limits_{n=1}^N e^{\langle z_{cm}, z_{cmn}^- \rangle / \tau}} \right].
\label{InfoNCE}
\end{aligned}
\end{equation}
It shows that we use contrastive learning to train our model on \(M\) anchor pixels from class \(c\), each paired with one positive sample and \(N\) negative samples. We denote the representations of the \(m\)-th anchor pixel and positive sample as \(z_{cm}\) and \(z_{c}^+\) where \(z\) is the output of the projection head. \(z_{cmn}^-\) denotes the \(n\)-th negative sample of the \(m\)-th anchor pixel. We measure the distance between features from different pixels by cosine similarity \(\langle \cdot, \cdot \rangle\) and use a temperature parameter \(\tau\) to learn from hard negatives. In the pixel-to-pixel methods, both positive and negative samples consist of pixel representations. In the pixel-to-prototype methods, both positive and negative samples are class prototypes. Contrary to these approaches, our technique employs class prototypes as positive and pixel representations as the negative instances. The merits of our approach have been elucidated in the introduction. Moreover, previous methods do not optimally exploit every pixel representation within the target domain. Conversely, our approach effectively integrates all target pixel representations into the contrastive learning during training. To employ the contrastive loss, we first need to select reliable target image pixels, and then utilize the source image annotations and target image pseudo-labels to determine and calculate the components required for contrastive learning. This matter will be elaborated upon in greater detail in the following sections, namely Section \ref{sec_III_B} and Section \ref{sec_III_C}.

To sum up, the aim of optimization is to lower all encompassed losses, delineated mathematically as:
\begin{equation}
\begin{aligned}
\mathcal{L} = \mathcal{L}_s + \lambda_ t\mathcal{L}_t + \lambda_c\mathcal{L}_c,
\label{overall loss}
\end{aligned}
\end{equation}
where $\lambda_t$ is the weight of target domain loss, and \(\lambda_c\) is the weight of contrastive loss. Moreover, the contrast loss is computed following a specified number of warmup iterations to maintain training stability.

\subsection{Selecting Reliable Target Image Pixels} \label{sec_III_B}
We employ per-pixel prediction entropy to filter high confident pixels for contrastive learning, which mitigates the negative influence of inaccurate pseudo-labels. Specifically, for pixel \(j\) of each \(i\)-th target image, \((P_t^i)^j\) is used to denote the softmax probabilities as output by the segmentation head of the teacher model. The entropy is calculated by:
\begin{equation}
\begin{aligned}
(E_t^i)^j = - \sum\limits_{c=0}^{C-1} (P_t^i)_c^j \log (P_t^i)_c^j,
\label{entropy}
\end{aligned}
\end{equation}
where \(C\) signifies the tally of distinct classes and \((P_t^i)_c^j\) corresponds to the \(c\)-th component of \((P_t^i)^j\). We consider pixels whose entropy is in the lowest \(\eta\%\) as confident pixels. \(\eta\%\) is a hyperparameter that will be determined through experiments.  Specifically, we set the threshold \(\gamma_t\) such that the fraction of elements in \(E_t^i\) that are lower than \(\gamma_t\) is equal to \(\eta\%\).

\subsection{Contrastive Learning} \label{sec_III_C}
\begin{figure*}[thbp]
	\centering
	\includegraphics[scale=1]{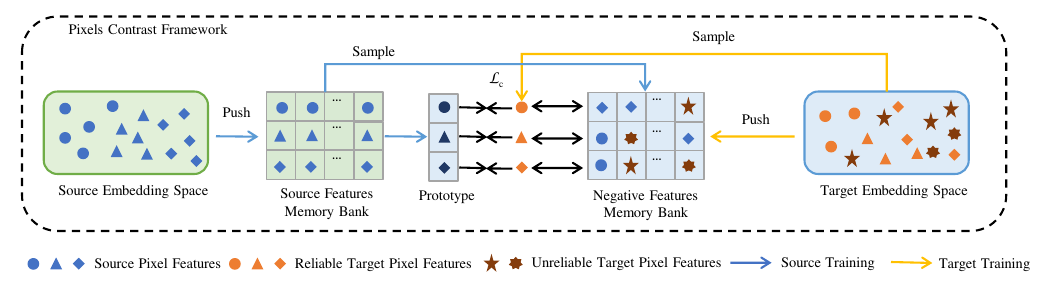}
	\doublespacing\caption{An overview of pixels contrast framework. Reliable target pixel features are sampled from the target embedding space, followed by decreasing the distance between these features and class prototypes while concurrently increasing the distance between them and negative features. To fully leverage the pixels from target images, unreliable target pixel features are introduced into the contrastive learning during training.}
	\label{contrastive_network} 
\end{figure*}

By employing contrastive learning, our proposed approach aims to improve the feature discrimination among different target classes, visualized in Fig. \ref{contrastive_network}. It is made up of three components: anchor pixels, positive samples, and negative samples. Note that for efficiency, anchor pixels and negative samples are sourced from certain sets, thus reducing computational costs. We describe these three components in the following paragraphs.

\noindent\textbf{Anchor Pixels.} 
In contrastive learning, anchor pixels serve as the starting points for constructing pairs of positive and negative samples and are utilized to guide feature learning. For every class within the current mini-batch, we select corresponding anchor pixels during training. Features from all candidate anchor pixels for category \(c\) are encapsulated within the set $\mathcal{A}_c$:
\begin{equation}
\begin{aligned}
\mathcal{A}_c = \{(z_t^{i})^j \mid (p_t^{i})^j = c \quad \cap\quad (E_t^i)^j < \gamma_t \},
\label{Anchor Pixels}
\end{aligned}
\end{equation}
where $(p_t^i)^j$ and \((z_t^{i})^j\) denote the pseudo-label and corresponding representation for pixel \(j\) within the target image \(i\), respectively.

\noindent\textbf{Positive Samples.} We utilize the prototype of each class as the positive sample in our method. To bring the prototype closer to the potential feature distribution center of class \(c\), we utilize not only the current mini-batch pixel representations but also a memory bank to compute the prototype. This memory bank can store a collection of pixel representations of class \(c\), allowing access to a broader range of representations for calculating the prototype:
\begin{equation}
\begin{aligned}
z_c^+ = \frac{1}{\vert\mathcal{M}_c\vert} \sum\limits_{z_c \in \mathcal{M}_c} z_c,
\label{Positive Samples}
\end{aligned}
\end{equation}
where \(\mathcal{M}_c\) denote all the source pixel representations for category \(c\) within the memory bank. Note that the prototype of category \(c\) is used for all anchor pixels assigned to the category \(c\).

\noindent\textbf{Negative Samples.} Negative Samples are included from two distinct realms: the source domain and the target domain. When considering \(i\)-th image of the source domain, a qualified negative sample of category \(c\) is defined as one that does not fall within category \(c\). Specifically, the collection of negative samples in source images for class \(c\) is
\begin{equation}
\begin{aligned}
\mathcal{N}_s^c = \{(z_s^{i})^j \mid (y_s^{i})^j \neq c\}.
\label{Source Negative Samples}
\end{aligned}
\end{equation}
On the other hand, with the objective of capitalizing on the information contained within all target pixels for improved discrimination, we operate under the assumption that the model can effectively distinguish between dissimilar classes while potentially encountering confusion primarily among a subset of classes that exhibit similarity. For instance, as depicted in Fig. \ref{Introduction_insight}(d), the pixel marked with an orange cross exhibits comparable probabilities for belonging to classes such as \(car\), \(bus\), and \(truck\) while the model exhibits a high level of confidence in rejecting classes like \(person\), \(vegetable\), and other similar classes. Consequently, we can utilize this pixel as a negative sample for classes like \(person\) or \(vegetable\). In this context, we define a qualified negative sample of category \(c\) concerning the \(i\)-th target image as one characterized by a low probability of being associated with category \(c\). To achieve this, we define the pixel-level category order $\mathcal{O}_{ij}$ as the sorted indices of $(P_t^i)^j$ in ascending order. It follows that \(\mathcal{O}_{ij}(\mathrm{argmax}(P_t^i)^j) = 0\) and \(\mathcal{O}_{ij}(\mathrm{argmin}(P_t^i)^j) = C - 1\), where \(C\) signifies the total count of distinct classes present.
Then the set of negative samples in target images for class \(c\) is 
\begin{equation}
\begin{aligned}
\mathcal{N}_t^c = \{(z_t^{i})^j \mid \mathbbm{1} [\mathcal{O}_{ij}(c) \geq r] = 1\},
\label{Target Negative Samples}
\end{aligned}
\end{equation}
where \(\mathbbm{1}\) is the notation for an indicator function and \(r\) is established as 4 to filter out features that have a low probability of belonging to class \(c\). However, the quantity of negatives is constrained by the mini-batch size, which hinders the contrastive representation learning  \cite{chen2020improved}. Therefore, we adopt large and external memories as a bank to store more negative samples of class \(c\), following \cite{chen2020improved}. Moreover, the stored negative samples are updated dynamically through a momentum-updated mechanism. 

After obtaining these components, we can calculate the contrastive loss according to Eq. \eqref{InfoNCE}. Algorithm \ref{algorithm_1} provides a structured overview of the principal stages of our technique.

\IncMargin{1em}
\begin{algorithm} \SetKwData{Left}{left}\SetKwData{This}{this}\SetKwData{Up}{up} \SetKwFunction{Union}{Union}\SetKwFunction{FindCompress}{FindCompress} \SetKwInOut{Input}{input}\SetKwInOut{Output}{output}
	
	\Input{Source data ${\mathcal{X}_s, \mathcal{Y}_s}$, target data $\mathcal{X}_t$, and hyperparameters.} 
	\Output{Final network weights $\Theta$.}
	\BlankLine 
	
	\emph{Initialize student $\Theta$ and teacher $\Theta'$ model parameters}\; 
	\For{$iteration\leftarrow 0$ \KwTo $total \quad iterations$}{ 
		\emph{Randomly sample source images $\mathcal{B}_s$ and target images $\mathcal{B}_t$}\;
		\emph{Compute target image predictions $p_t$ using teacher model}\;
		\If{$iteration$ $>$ $warmup \quad iterations$}
		{\emph{Separate pixel-level representation of both domains in the embedding space}\;
			\emph{Get anchor pixels $z_{cm}$ from $\mathcal{A}_c$}\;
			\emph{Get negative samples $z_{cmn}^-$ from $\mathcal{N}_s^c$ and $\mathcal{N}_t^c$}\;
			\emph{Get positive samples $z_c^+$ based on Eq. \eqref{Positive Samples}}\;
			\emph{Compute contrastive loss $\mathcal{L}_c$ based on Eq. \eqref{InfoNCE}}\;}
		\emph{Compute loss for the mini-batch $\mathcal{L}$ based on Eq. \eqref{overall loss} and backpropagation}\;
		\emph{Update $\Theta$ and $\Theta'$}.}
	\caption{PGPC}
	\label{algorithm_1} 
\end{algorithm}
\DecMargin{1em} 

\section{Experiments}

This section elaborates on the experimental setup, encompassing datasets, metrics for evaluation, utilized architectures, and implementation details. Subsequently, it exhibits the outcomes of our proposed approach and demonstrates its effectiveness.
\subsection{Experimental Details}
\noindent\textbf{Datasets.} The Cityscapes dataset \cite{cityscapes} encompasses a collection of 5,000 images, each with pixel-wise semantic annotations, alongside 20,000 images featuring broader semantic annotations. These visual data are collected from 50 different real-world cities. The target domain is composed of 2975 unlabeled images from the Cityscapes training dataset. To gauge the efficacy of our model, its performance is meticulously assessed using a separate set of 500 images from the Cityscapes validation dataset, which benefits from comprehensive labeling.

The GTA5 dataset \cite{GTA} has 24,966 computer-generated images originating from a virtual gaming engine. The network is trained on 19 classes that are common between GTA5 and Cityscapes, and the rest of the classes are ignored.

The SYNTHIA dataset \cite{synthia}, a virtual compilation of imagery, includes 9,400 urban scenes with high-quality masks. It is used to train the network on 13 classes that correspond to those found in the Cityscapes dataset.

\noindent\textbf{Evaluation.} The mean intersection-over-union (mIoU) metric, a standard in UDA for semantic segmentation, is utilized to assess the efficacy of semantic segmentation networks. 

\noindent\textbf{Architectures.} The implementation of our approach follows the recent advanced UDA setting and is based on the DAFormer \cite{daformer}. We also adopt HRDA \cite{hrda} to further substantiate the efficacy of our approach. Specifically, we utilize the MiT-B5 encoder \cite{segformer} pretrained on ImageNet-1k \cite{imagenet} to generate a feature pyramid with $\mathcal{F}_d = [64, 128, 320, 512]$. The features are passed on a segmentation head and a projection head. The segmentation head aligns with the configuration of the decoder in DAFormer \cite{daformer} that uses a multitude of multiple $3 \times 3$ depthwise separable convolutions operating in parallel and characterized by a spectrum of dilation rates. Concurrently, the projection head facilitates the transformation of high-dimensional pixel embeddings into a feature vector of 256 dimensions, normalized through the $l_2$ scheme. Furthermore, in the context of HRDA \cite{hrda}, an additional scale attention head is implemented, which parallels the streamlined MLP-based decoder of the SegFormer model \cite{segformer}, with an embedding dimension also set to 256.

\noindent\textbf{Implementation details.} Experiments are conducted using PyTorch and are based on the mmsegmentation toolbox \cite{mmseg}. We adopt the training approach as DAFormer \cite{daformer} and HRDA \cite{hrda} and use AdamW \cite{AdamW} optimizer, characterized by betas \((0.9, 0.999)\) and a weight decay of $0.01$. Moreover, the learning rate of the encoder is configured to $6 \times 10 ^{-5}$, while $6 \times 10^{-4}$ is the designated learning rate for the segmentation head and the projection head. For self-training, we keep $\alpha = 0.999$ as the EMA weight update parameter and apply three training strategies: linear learning rate warmup policy, rare class sampling, and feature distance regularization. Throughout the training phase, we train our semantic segmentation network for 60k iterations. Moreover, for an equitable comparison, we set $\kappa$ to 0.968 following DAFormer \cite{daformer}. For the contrastive learning, we selected the values of the warmup iterations and the temperature parameter \(\tau\) based on the literature and our preliminary experiments. Specifically, we used 20k warmup iterations and \(\tau = 0.5\). We tuned the other hyperparameters empirically to optimize the performance of our model.

\subsection{Experimental Results}
\begin{figure*}[htbp]
	\centering
	\includegraphics[scale=1]{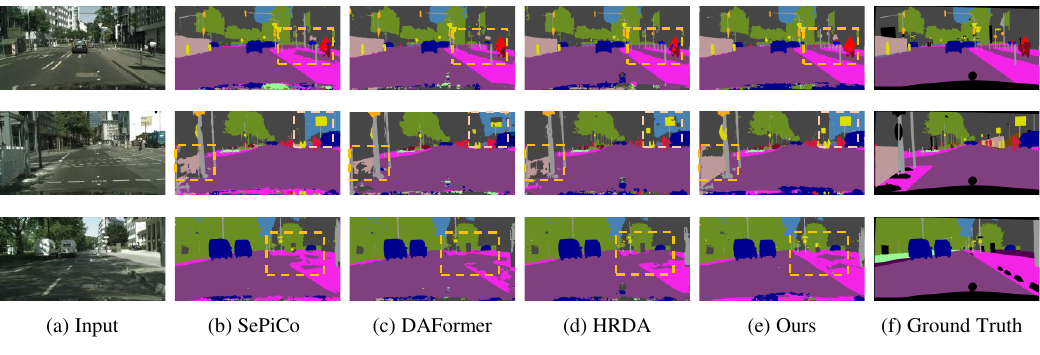}
	\doublespacing\caption{Qualitative examples of semantic segmentation on the GTA5 to Cityscapess. (a) and (f) are the images and the ground truth labels from the Cityscapes validation dataset. (b), (c), and (d) are the segmentation predictions of SePiCo \cite{sepico}, DAFormer \cite{daformer}, and HRDA \cite{hrda}. (e) are the segmentation outputs of our method. We emphasize areas of superior performance by our method with highlighted colored boxes.}
	\label{effectiveness} 
\end{figure*}

We execute an in-depth examination of our approach against the advanced approaches \cite{daformer, sepico, hrda} within two exemplary virtual-to-real adaptation tasks: GTA5 to Cityscapes, SYNTHIA to Cityscapes, Cityscapes to DarkZurich and Citycapes to Adverse Conditions Dataset
with Correspondences (ACDC). The quantitative results are presented in Tables \ref{tab: table1} and \ref{tab: table2}. Qualitative results of the segmentation effectiveness are illustrated in Figure \ref{effectiveness}. Our method is validated to secure outstanding segmentation performance in each scenario, using different segmentation models such as DAFormer \cite{daformer} and HRDA \cite{hrda}.

\noindent\textbf{GTA5 to Cityscapes.} TABLE \ref{tab: table1} exhibits the adaptation results on the task of GTA5 to Cityscapes, where we compare our approach with the advanced UDA approaches \cite{sepico, cluda, prototypical_contrast, ProDA}. Optimal results are denoted with bold formatting. Our method achieves a leading result of $71.8$ mIoU, which significantly surpasses DAFormer by $+3.5$ mIoU. This indicates that our method can effectively leverage contrastive learning and pixel-level alignment to improve the domain adaptation performance. Moreover, when we apply our method to HRDA, we further boost the mIoU by $+1.2$ mIoU, demonstrating the scalability and robustness of our method to different segmentation models.

\noindent\textbf{SYNTHIA to Cityscapes.} TABLE \ref{tab: table2} showcases the adaptation outcomes on SYNTHIA to Cityscapes. Our approach yields a mIoU score of $67.7$, rivaling the performance of advanced approaches such as SePiCo \cite{sepico} and CLUDA \cite{cluda}. In addition, our method outperforms them in some challenging classes. This shows that our approach successfully extends its applicability to different virtual-to-real adaptation scenarios and captures the semantic features of the target domain.

\noindent\textbf{Citycapes to DarkZurich.} TABLE \ref{tab: cs2dzur} shows the outcomes of Citycapes to DarkZurich. Our method achieved the best performance of $61.2$ mIoU, improving by $1.1$ mIoU compared to MIC (HRDA) \cite{MIC}. Moreover, our method demonstrates better domain adaptation capabilities, for example, in the class \(bus\), our method improved by 35.1\% compared to the previous best approach, MIC (HRDA) \cite{MIC}. 

\noindent\textbf{Citycapes to ACDC.} We test different models on Citycapes to ACDC task, and the result is shown in Table \ref{tab: cs2acdc}. Our method achieve $71.4$ mIoU, which is an increase of $1.0$ mIoU compared to the best past method, MIC (HRDA) \cite{MIC}. In harsh environmental conditions, our method still performed at its best, indicating that our approach has a certain degree of robustness against interference.

\noindent\textbf{Qualitative results.} Fig. \ref{effectiveness} illustrates some segmentation examples and compares them with the ground truth and the predictions of SePiCo \cite{sepico}, DAFormer \cite{daformer}, and HRDA \cite{hrda}. We use colored boxes to indicate the regions where our method can segment minor categories better than the other methods. The proficiency of our method in generating precise segmentation is evident for minority classes, with sharper and more accurate boundaries between categories.

\noindent\textbf{The speed and complexity.} Our method integrates contrastive learning with DAFormer \cite{daformer}, which does not increase model complexity
 but increases training time and inference time. Table \ref{tab: Complexity} presents the training speed and inference speed of various methods. 
 In the Cityscapes to Darkzurich task, training a single image with PGPC + DAFormer takes 33\% longer compared to DAFormer, 
 while PGPC + HRDA shows a 33\% increase in training time relative to HRDA \cite{hrda}. 
 The inference speed of PGPC + HRDA is not different from HRDA\cite{hrda}, but PGPC + DAFormer takes 3.6 times as long as DAFormer.

\settowidth\rotheadsize{sidewalk}
\noindent
\begin{table*}[thbp]
	\doublespacing\caption{The mIoU $\%$ scores of semantic segmentation models trained on GTA5 and evaluated on the Cityscapes validation dataset. The preeminent scores within columns are highlighted through bold text. \label{tab: table1}}
	\centering\small
	\begin{tabularx}{\linewidth}{l|*{19}{b{0.32cm}}|l}
		\toprule
		\multicolumn{21}{c}{GTA5 to Cityscapes}\\
		\midrule
		Method& \rothead{road}& \rothead{sidewalk}& \rothead{building}& \rothead{wall}& \rothead{fence}& \rothead{pole}& \rothead{light}& \rothead{sign}& \rothead{veg}& \rothead{terrain}& \rothead{sky}& \rothead{person}& \rothead{rider}& \rothead{car}& \rothead{truck}& \rothead{bus}& \rothead{train}& \rothead{mbike}& \rothead{bike}& mIoU \\
		\midrule
		IAST\cite{IAST}& 94.1& 58.8& 85.4& 39.7& 29.2& 25.1& 43.1& 34.2& 84.8& 34.6& 88.7& 62.7& 30.3& 87.6& 42.3& 50.3& 24.7& 35.2& 40.2& 52.2\\
		UncerDA\cite{uncertainty_aware}& 90.5& 38.7& 86.5& 41.1& 32.9& 40.5& 48.2& 42.1& 86.5& 36.8& 84.2& 64.5& 38.1& 87.2& 34.8& 50.4& 0.2& 41.8& 54.6& 52.6\\ 
		DACS\cite{dacs}& 89.9& 39.7& 87.9& 30.7& 39.5& 38.5& 46.4& 52.8& 88.0& 44.0& 88.8& 67.2& 35.8& 84.5& 45.7& 50.2& 0.0& 27.3& 34.0& 52.1\\
		SAC\cite{DA_SAC}& 90.4& 53.9& 86.6& 42.4& 27.3& 45.1& 48.5& 42.7& 87.4& 40.1& 86.1& 67.5& 29.7& 88.5& 49.1&  54.6& 9.8& 26.6& 45.3& 53.8\\
		ProCA\cite{prototypical_contrast}& 91.9& 48.4& 87.3& 41.5& 31.8& 41.9& 47.9& 36.7& 86.5& 42.3& 84.7& 68.4& 43.1& 88.1& 39.6& 48.8& 40.6& 43.6& 56.9& 56.3\\
		BCL\cite{lee2022bi}& 93.5& 60.2& 88.1& 31.1& 37.0& 41.9& 54.7& 37.8& 89.9& 45.5& 89.9& 72.7& 38.2& 90.7& 34.3& 53.2& 4.4& 47.2& 58.5& 57.1\\
		ProDA\cite{ProDA}& 87.8& 56.0& 79.7& 46.3& 44.8& 45.6& 53.5& 53.5& 88.6& 45.2& 82.1& 70.7& 39.2& 88.8& 45.5& 59.4& 1.0& 48.9& 56.4& 57.5\\
		CaCo\cite{huang2022category}& 93.8& 64.1& 85.7& 43.7& 42.2& 46.1& 50.1& 54.0& 88.7& 47.0& 86.5& 68.1& 2.9& 88.0& 43.4& 60.1& 31.5& 46.1& 60.9& 58.0\\
		DAFormer\cite{daformer}& 95.7& 70.2& 89.4& 53.5& 48.1& 49.6& 55.8& 59.4& 89.9& 47.9& 92.5& 72.2& 44.7& 92.3& 74.5& 78.2& 65.1& 55.9& 61.8& 68.3\\
		{\footnotesize CLUDA+DAFormer\cite{cluda}}& \textbf{97.5}& 78.8& 88.8& 60.8& 52.0& 47.1& 51.9& 50.3& 89.7& 51.0& 94.0& 71.0& 48.6& 93.1& 82.0& 84.1& 71.4& 58.9& 60.7& 70.1\\
		SePiCo\cite{sepico}& 96.9& 76.7& 89.7& 55.5& 49.5& 53.2& 60.0& 64.5& 90.2& 50.3& 90.8& 74.5& 44.2& 93.3& 77.0& 79.5& 63.6& 61.0& 65.3& 70.3\\ 
		MIC(DAFormer)\cite{MIC}& 96.7& 75.0& 90.0& 58.2& 50.4& 51.1& 56.7& 62.1& 90.2& 51.3& 92.9& 72.4& 47.1& 92.8& 78.9& 83.4& 75.6& 54.2& 62.6& 70.6\\ 
		HRDA\cite{hrda}& 96.4& 74.4& 91.0& 61.6& 51.5& 57.1& 63.9& 69.3& 91.3& 48.4& 94.2& 79.0& 52.9& 93.9& 84.1& 85.7& 75.9& 63.9& 67.5& 73.8\\
		CLUDA+HRDA\cite{cluda}& 97.1& 78.0& 91.0& 60.3& 55.3& 56.3& 64.3& 71.5& 91.2& 51.1& 94.7& 78.4& 52.9& 94.5& 82.8& 86.5& 73.0& 64.2& 69.7& 74.4\\
		PDA\cite{PDA}& 97.8& 78.1& 92.4& \textbf{70.8}& 53.1& 58.5& 58.2& 67.1& 92.0& \textbf{54.4}& 94.8& \textbf{81.4}& 54.5& 92.9& \textbf{86.9}& 87.1& 78.6& \textbf{68.1}& 69.7& 75.6\\
		MIC(HRDA)\cite{MIC}& 97.4& 80.1& 91.7& 61.2& 56.9& \textbf{59.7}& \textbf{66.0}& 71.3& 91.7& 51.4& 94.3& 79.8& 56.1& 94.6& 85.4& \textbf{90.3}& \textbf{80.4}& 64.5& 68.5& 75.9\\ 
		\midrule
		PGPC+ProCA& 93.1& 51.6& 89.4& 44.7& 40.5& 49.2& 52.4& 45.6& 88.3& 48.7& 86.9& 70.2& 40.2& 89.2& 70.9& 50.7& 48.6& 46.7& 58.2& 61.3\\
		PGPC+DAFormer& 97.1& 77.4& 90.2& 56.8& 51.7& 53.9& 61.2& 66.8& 90.4& 50.0& 93.0& 74.6& 41.5& 93.4& 80.5& 85.4& 78.6& 59.9& 62.5& 71.8\\
		{\footnotesize PGPC+MIC(DAFormer)}& 96.4& 77.3& 90.7& 55.2& 54.8& 56.2& 60.2& 67.2& 90.9& 51.0& 86.7& 75.4& 49.6& 93.8& 83.5& 83.6& 76.3& 60.2& 65.0& 72.3\\
		PGPC+HRDA& 97.2& 75.4& 92.4& 64.1& 58.4& 58.7& 64.4& 71.4& 91.7& 53.4& 94.7& 76.5& 53.0& 94.4& 86.1& 85.2& 75.0& 65.4& 68.5& 75.0\\
		{PGPC+MIC(HRDA)}& \textbf{97.5}& \textbf{80.7}& \textbf{92.8}& 63.7& \textbf{59.3}& 58.1& 65.3& \textbf{71.9}& \textbf{92.4}& 53.0& \textbf{95.2}& 79.5& \textbf{58.8}& \textbf{95.1}& 86.2& 89.3& 79.2& 64.9& \textbf{70.0}& \textbf{76.5}\\
		\bottomrule
	\end{tabularx}
\end{table*}

\begin{table*}[htbp]
	\doublespacing\caption{The mIoU $\%$ scores of semantic segmentation models trained on SYNTHIA and evaluated on the Cityscapes validation dataset. The preeminent scores within columns are highlighted through bold text. \label{tab: table2}}
	\centering\small
	\begin{tabular}{l|*{16}{b{0.3cm}}|c}
		\toprule
		\multicolumn{18}{c}{SYNTHIA to Cityscapes}\\
		\midrule
		Method& \rothead{road}& \rothead{sidewalk}& \rothead{building}& \rothead{wall}& \rothead{fence}& \rothead{pole}& \rothead{light}& \rothead{sign}& \rothead{veg}& \rothead{sky}& \rothead{person}& \rothead{rider}& \rothead{car}& \rothead{bus}& \rothead{mbike}& \rothead{bike}& mIoU \\
		\midrule
		UncerDA\cite{uncertainty_aware}& 79.4& 34.6& 83.5& 19.3& 2.8& 35.3& 32.1& 26.9& 78.8& 79.6& 66.6& 30.3& 86.1& 36.6& 19.5& 56.9& 48.0\\
		IAST\cite{IAST}& 81.9& 41.5& 83.3& 17.7& 4.6& 32.3& 30.9& 28.8& 83.4& 85.0& 65.5& 30.8& 86.5& 38.2& 33.1& 52.7& 49.8\\
		SAC\cite{DA_SAC}& 89.3& 47.2& 85.5& 26.5& 1.3& 43.0& 45.5& 32.0& 87.1& 89.3& 63.6& 25.4& 86.9& 35.6&  30.4& 53.0& 59.3\\
		ProCA\cite{prototypical_contrast}& \textbf{90.5}& 52.1& 84.6& 29.2& 3.3& 40.3& 37.4& 27.3& 86.4& 85.9& 69.8& 28.7& 88.7& 53.7& 14.8& 54.8& 53.0\\
		ProDA\cite{ProDA}& 87.8& 45.7& 84.6& 37.1& 0.6& 44.0& 54.6& 37.0& 88.1& 84.4& 74.2& 24.3& 88.2& 51.1& 40.5& 45.6& 55.5\\
		BCL\cite{lee2022bi}& 83.8& 42.2& 85.3& 16.4& 5.7& 43.1& 48.3& 30.2& 89.3& 92.1& 68.2& 43.1& 89.7& 47.2& 42.2& 54.2& 55.6\\
		DAFormer\cite{daformer}& 84.5& 40.7& 88.4& 41.5& 4.5& 53.1& 55.0& 54.6& 86.0& 89.8& 73.2& 48.2& 87.2& 53.2& 53.9& 61.7& 60.9\\
		CLUDA+DAFormer\cite{cluda}& 87.4& 44.8& 86.5& 47.9& 8.7& 49.8& 44.5& 52.7& 85.6& 89.2& 74.4& 50.2& 86.9& 65.3& 56.9& 57.1& 61.7\\
		MIC(DAFormer)\cite{MIC}& 83.0& 40.9& 88.2& 37.6& 9.0& 52.4& 56.0& 56.5& 87.6& 93.4& 74.2& 51.4& 87.1& 59.6& 57.9& 61.2& 62.2\\
		SePiCo\cite{sepico}& 87.0& 52.6& 88.5& 40.6& 10.6& 49.8& 57.0& 55.4& 86.8& 86.2& 75.4& 52.7& \textbf{92.4}& \textbf{78.9}& 53.0& 62.6& 64.3\\
		HRDA\cite{hrda}& 85.2& 47.7& 88.8& 49.5& 4.8& 57.2& 65.7& 60.9& 85.3& 92.9& 79.4& 52.8& 89.0& 64.7& 63.9& 64.9& 65.8\\
		PDA\cite{PDA}& 90.2& 52.1& 88.3& \textbf{50.5}& \textbf{12.7}& 54.1& 65.2& 61.4& 86.2& 90.8& 80.4& 55.2& 89.5& 65.1& 64.1& \textbf{68.9}& 67.1\\  
		CLUDA+HRDA\cite{cluda}& 87.7& 46.9& \textbf{90.2}& 49.0& 7.9& 59.5& 66.9& 58.5& 88.3& \textbf{94.6}& 80.1& 57.1& 89.8& 68.2& 65.5& 65.8& 67.2\\
		MIC(HRDA)\cite{MIC}& 86.6& 50.5& 89.3& 47.9& 7.8& 59.4& 66.7& \textbf{63.4}& 87.1& \textbf{94.6}& \textbf{81.0}& 58.9& 90.1& 61.9& \textbf{67.1}& 64.3& 67.3\\
		\midrule
		PGPC+ProCA& 87.4& \textbf{54.0}& 86.1& 40.8& 5.4& 47.3& 46.7& 30.4& 89.4& 88.1& 70.8& 35.7& 85.4& 57.6& 40.4& 58.9& 57.8\\
		PGPC+DAFormer& 88.2& 51.3& 87.9& 49.5& 5.8& 53.1& 60.8& 48.4& 87.6& 92.7& 76.4& 49.7& 87.2& 65.4& 57.3& 58.5& 63.7\\
		{\footnotesize PGPC+MIC(DAFormer)}& 88.4& 52.4& 87.6& 49.2& 6.4& 54.5& 61.0& 49.5& 88.4& 92.1& 77.2& 50.0& 87.6& 66.1& 56.9& 60.1& 64.2\\
		PGPC+HRDA& 89.2& 48.8& 89.4& 49.1& 7.4& 61.2& 67.0& 61.2& 89.1& 94.2& 80.4& 58.6& 91.1& 69.1& 64.9& 62.2& 67.7\\
		{PGPC+MIC(HRDA)}& 89.4& 53.5& 89.4& 49.7& 6.7& \textbf{63.1}& \textbf{68.4}& 60.5& \textbf{89.7}& 93.9& 80.8& \textbf{59.2}& 91.4& 68.5& 65.4& 64.9& \textbf{68.4}\\
		\bottomrule
	\end{tabular}
\end{table*}

\settowidth\rotheadsize{sidewalk}
\noindent
\begin{table*}[thbp]
	\doublespacing\caption{The mIoU $\%$ scores of semantic segmentation models trained on Cityscapes and evaluated on the DarkZurich test dataset. The preeminent scores within columns are highlighted through bold text. \label{tab: cs2dzur}}
	\centering\small
	\begin{tabularx}{\linewidth}{l|*{19}{b{0.32cm}}|l}
		\toprule
		\multicolumn{21}{c}{Citycapes to DarkZurich}\\
		\midrule
		Method& \rothead{road}& \rothead{sidewalk}& \rothead{building}& \rothead{wall}& \rothead{fence}& \rothead{pole}& \rothead{light}& \rothead{sign}& \rothead{veg}& \rothead{terrain}& \rothead{sky}& \rothead{person}& \rothead{rider}& \rothead{car}& \rothead{truck}& \rothead{bus}& \rothead{train}& \rothead{mbike}& \rothead{bike}& mIoU \\
		\midrule
		ADVENT\cite{advent}& 85.8& 37.9& 55.5& 27.7& 14.5& 23.1& 14.0& 21.1& 32.1& 8.7& 2.0& 39.9& 16.6& 64.0& 13.8& 0.0& 58.8& 28.5& 27.0& 29.7\\
		GCMA\cite{gcma}& 81.7& 46.9& 58.8& 22.0& 20.0& 41.2& 40.5& 41.6& 64.8& 31.0& 32.1& 53.5& 47.5& 75.5& 39.2& 0.0& 49.6& 30.7& 21.0& 42.0\\
		MGCDA\cite{mgcda}& 80.3& 49.3& 66.2& 7.8& 11.0& 41.4& 38.9& 39.0& 64.1& 18.0& 55.8& 52.1& 53.5& 74.7& 66.0& 0.0& 37.5& 29.1& 22.7& 42.5\\
		DANNet\cite{dannet}& 90.0& 54.0& 74.8& 41.0& 21.1& 25.0& 26.8& 30.2& \textbf{72.0}& 26.2& 84.0& 47.0& 33.9& 68.2& 19.0& 0.3& 66.4& 38.3& 23.6& 44.3\\
		DAFormer\cite{daformer}& 93.5& 65.5& 73.3& 39.4& 19.2& 53.3& 44.1& 44.0& 59.5& 34.5& 66.6& 53.4& 52.7& 82.1& 52.7& 9.5& 89.3& 50.5& 38.5& 53.8\\
		HRDA\cite{hrda}& 90.4& 56.3& 72.0& 39.5& 19.5& 57.8& 52.7& 43.1& 59.3& 29.1& \textbf{70.5}& 60.0& 58.6& \textbf{84.0}& 75.5& 11.2& 90.5& 51.6& 40.9& 55.9\\
		MIC(HRDA)\cite{MIC}& \textbf{94.8}& \textbf{75.0}& 84.0& 55.1& \textbf{28.4}& 62.0& 35.5& 52.6& 59.2& 46.8& 70.0& 65.2& 61.7& 82.1& 64.2& 18.5& 91.3& 52.6& 44.0& 60.2\\ 
		\midrule
		PGPC+DAFormer& 90.4& 54.3& 81.6& \textbf{56.3}& 24.6& 60.0& 40.0& 51.7& 42.5& 40.0& 39.4& 62.8& 61.7& 73.3& 72.2& 4.3& 91.9& 49.7& 43.9& 54.7\\
		{\footnotesize PGPC+MIC(DAFormer)}& 92.5& 59.5& 79.3& 53.5& 24.5& 60.7& 42.6& 53.4& 42.3& 43.9& 39.3& 63.8& 62.6& 73.3& 71.5& 6.0& \textbf{93.3}& 52.17& 43.1& 55.7\\
		{PGPC+HRDA}& 90.9& 57.6& 79.4& 45.3& 21.6& 60.5& 32.2& 50.6& 56.2& 39.8& 66.5& 63.7& 53.7& 83.6& 70.1& \textbf{25.0}& 92.1& 56.8& 41.2& 57.2\\
		PGPC+MIC(HRDA)& 94.5& 65.7& \textbf{85.4}& 54.7& 27.5& \textbf{62.3}& \textbf{56.0}& \textbf{54.8}& 58.0& \textbf{47.5}& 68.8& \textbf{66.2}& \textbf{65.6}& 75.5& \textbf{79.4}& 4.5& 93.2& \textbf{61.2}& \textbf{45.6}& \textbf{61.3}\\
		\bottomrule
	\end{tabularx}
\end{table*}

\settowidth\rotheadsize{sidewalk}
\noindent
\begin{table*}[thbp]
	\doublespacing\caption{The mIoU $\%$ scores of semantic segmentation models trained on Cityscapes and evaluated on the ACDC test dataset. The preeminent scores within columns are highlighted through bold text. \label{tab: cs2acdc}}
	\centering\small
	\begin{tabularx}{\linewidth}{l|*{19}{b{0.32cm}}|l}
		\toprule
		\multicolumn{21}{c}{Citycapes to ACDC}\\
		\midrule
		Method& \rothead{road}& \rothead{sidewalk}& \rothead{building}& \rothead{wall}& \rothead{fence}& \rothead{pole}& \rothead{light}& \rothead{sign}& \rothead{veg}& \rothead{terrain}& \rothead{sky}& \rothead{person}& \rothead{rider}& \rothead{car}& \rothead{truck}& \rothead{bus}& \rothead{train}& \rothead{mbike}& \rothead{bike}& mIoU \\
		\midrule
		ADVENT\cite{advent}& 72.9& 14.3& 40.5& 16.6& 21.2& 9.3& 17.4& 21.2& 63.8& 23.8& 18.3& 32.6& 19.5& 69.5& 36.2& 34.5& 46.2& 26.9& 36.1& 32.7\\
		GCMA\cite{gcma}& 79.7& 48.7& 71.5& 21.6& 29.9& 42.5& 56.7& 57.7& 75.8& 39.5& 87.2& 57.4& 29.7& 80.6& 44.9& 46.2& 62.0& 37.2& 46.5& 53.4\\\
		MGCDA\cite{mgcda}& 73.4& 28.7& 69.9& 19.3& 26.3& 36.8& 53.0& 53.3&  75.4& 32.0& 84.6& 51.0& 26.1& 77.6& 43.2& 45.9& 53.9& 32.7& 41.5& 48.7\\
		DANNet\cite{dannet}& 84.3& 54.2& 77.6& 38.0& 30.0& 18.9& 41.6& 35.2& 71.3& 39.4& 86.6& 48.7& 29.2& 76.2& 41.6& 43.0& 58.6& 32.6& 43.9& 50.0\\
		DAFormer\cite{daformer}& 58.4& 51.3& 84.0& 42.7& 35.1& 50.7& 30.0& 57.0& 74.8& 52.8& 51.3& 58.3& 32.6& 82.7& 58.3& 54.9& 82.4& 44.1& 50.7& 55.4\\
		HRDA\cite{hrda}& 88.3& 57.9& 88.1& 55.2& 36.7& 56.3& 62.9& 65.3& 74.2& 57.7& 85.9& 68.8& 45.7& 88.5& \textbf{76.4}& 82.4& 87.7& 52.7& 60.4& 68.0\\ 
		MIC(HRDA)\cite{MIC}& 90.8& \textbf{67.1}& 89.2& 54.5& 40.5& 57.2& 62.0& 68.4& 76.3& \textbf{61.8}& 87.0& \textbf{71.3}& \textbf{49.4}& 89.7& 75.7& 86.8& \textbf{89.1}& 56.9& 63.0& 70.4\\
		\midrule
		PGPC+DAFormer& 83.2& 44.6& 83.8& 42.5& 30.8& 48.0& 39.0& 55.5& 68.0& 52.2& 80.5& 57.9& 24.0& 76.0& 51.6& 66.0& 84.4& 39.1& 46.7& 56.5 \\
		{\footnotesize PGPC+MIC(DAFormer)}& \textbf{91.6}& 68.7& 86.2& 55.7& 34.3& 54.8& 50.6& 34.1& 73.1& 56.9& 84.5& 65.0& 46.4& 86.1& 67.8& 71.4& 84.2& 52.7& 55.9& 64.2\\ 
		PGPC+HRDA& 72.0& 58.6& 88.3& 45.9& 35.2& 52.7& 61.0& 63.5& 77.7& 58.0& 72.1& 66.3& 47.0& 86.7& 67.6& 80.7& 86.4& 48.8& 58.3& 64.6\\
		{PGPC+MIC(HRDA)}& 85.9& 60.1& \textbf{93.1}& \textbf{63.2}& \textbf{44.0}& \textbf{62.9}& \textbf{71.7}& \textbf{69.1}& \textbf{92.4}& 41.9& \textbf{96.9}& 69.8& 30.5& \textbf{91.3}& 81.7& \textbf{89.3}& 89.0& \textbf{57.4}& \textbf{66.5}& \textbf{71.4}\\
		\bottomrule
	\end{tabularx}
\end{table*}

\begin{table}[htbp]
	\doublespacing\caption{The speed of different models. ($frames/s$) \label{tab: Complexity}}
	\centering\small
	\begin{tabular}{l|ccccc}
		\toprule
		Method& Training speed& Inference speed \\
		\midrule
		Daformer \cite{daformer}&  0.8& 8.9\\
		HRDA \cite{hrda}&  0.4& 0.5\\
		PGPC+DAFormer&  0.6& 2.5\\ 
		PGPC+HRDA&  0.3& 0.5\\	
		\bottomrule
	\end{tabular}
\end{table}

\subsection{Ablation Studies and Further Discussion}
\begin{figure}[h]
	\centering
	\includegraphics[scale=1.5]{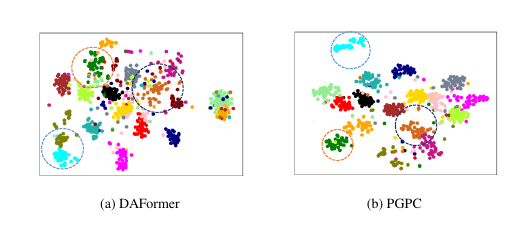}
	\doublespacing\caption{The visualization of our method (PGPC) and DAFormer \cite{daformer}. Different colors represent distinct categories, and these colors are consistent with the category label colors. We highlight some categories where our method performs better than DAFormer with colored circles.}
	\label{TSNE} 
\end{figure}

\noindent\textbf{Effectiveness of the contrastive learning.} To evaluate the influence of contrastive learning on the domain adaptation task of GTA5 to Cityscapes, we conduct our experiments using the same setup and hyperparameters as the DAFormer \cite{daformer}. Table \ref{tab: table3} shows that contrastive learning can improve the mIoU by $+3.5$ compared with the baseline. Moreover, our experimental approach includes assessing how different selections of positive and negative samples affect the performance. Table \ref{tab: table3} reveals that using target negative pixels contributes to a significant increase in mIoU. This affirms our viewpoint that the comprehensive potential of target images can be fully realized through the optimization of individual pixel utilization. In addition, using source prototypes rather than target prototypes as positive samples can better narrow the domain gap, thereby achieving improved results. From Table \ref{tab: table3} and Table \ref{tab: table1}, even without incorporating the use of all target pixels, our proposed framework (70.8 mIoU) still outperforms pixel-to-pixel (CLUDA: 70.1 mIoU) and pixel-to-prototype (SePiCo: 70.3 mIoU) methods, as elaborated in the introduction section. In an effort to delineate the ramifications of our method in a more illustrative and understandable fashion, we present the visualization \cite{tsne} of learned feature representations for DAFormer \cite{daformer} and PGPC, as shown in Fig. \ref{TSNE}. Through the visualization, we are able to discern that the clusters formed by the representations of PGPC demonstrate enhanced clarity when compared to those generated by DAFormer \cite{daformer}. We use circles of different colors in the Fig. \ref{TSNE} to further illustrate the performance of our approach. We can observe that the feature points in our method converge more closely while there are many scattered points in DAFormer \cite{daformer} by comparing the feature points within these circles. Nevertheless, it is also discernible that certain shortcomings persist in the results. We attribute these shortcomings to erroneous pseudo-labeling, a challenge commonly encountered in unsupervised self-training. Despite our best efforts to minimize the effect of incorrect pseudo-labels, it is currently unavoidable. Eliminating the influence stemming from incorrect pseudo-labels remains a pivotal focus for future research.

\begin{table}[htbp]
	\doublespacing\caption{Effectiveness of the contrastive learning on GTA5 to Cityscapes. SP represents the use of source prototypes as the positive samples. TP means the use of target prototypes as positive samples. SN means the source negative pixels while TN means the target negative pixels. Note that SP and TP cannot be used simultaneously as positive samples. \label{tab: table3}}
	\centering\small
	\begin{tabular}{l|cc|cc|c|c}
		\toprule
		Method& SP& TP& SN& TN& mIoU& $\triangle$mIoU\\
		\midrule
		\multirow{6}{*}{PGPC}& \checkmark& & \checkmark& \checkmark& 71.8 & +3.5\\
		& \checkmark & & \checkmark& & 70.8&  +2.5\\
		& \checkmark & & & \checkmark& 70.0&  +1.7\\
		& & \checkmark& \checkmark& \checkmark& 71.0&  +2.7\\
		& & \checkmark& \checkmark& & 70.1&  +1.8\\
		& & \checkmark& & \checkmark& 69.8&  +1.5\\
		\midrule
		DAFormer& & & & & 68.3&  $-$\\
		\bottomrule
	\end{tabular}
\end{table}

\begin{figure}[h]
	\centering
	\includegraphics[scale=1.5]{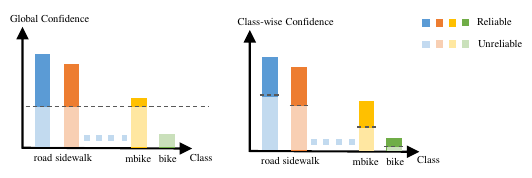}
	\doublespacing\caption{The visualization of global confidence and class-wise confidence methods. Class-wise confidence produces unreliable pseudo-labels for minority classes compared to global confidence, which amplifies noise in contrastive learning and ultimately reduces overall performance.}
	\label{thresholding_method} 
\end{figure}

\begin{table}[htbp]
	\doublespacing\caption{Analysis of the different methods of selecting reliable pixels on GTA5 to Cityscapes. \label{tab: thresholding method}}
	\centering\small
	\begin{tabular}{l|c|c|c|c}
		\toprule
		\multicolumn{1}{c|}{\multirow{2}{*}{Method}}& \multicolumn{2}{c|}{Global} & \multicolumn{2}{c}{Class-wise} \\	\cline{2-5}
		& Entropy & Confidence& Entropy& Confidence\\
		\midrule
		mIoU& \textbf{71.8}& 70.9& 70.9&  70.3\\
		\bottomrule
	\end{tabular}
\end{table}

\noindent\textbf{Different methods of selecting reliable pixels.} In our method, we employ global entropy to select reliable target image pixels, which is a widely used approach \cite{pixelIntraDA}. Additionally, there are three other common methods for selecting reliable pixels. One is to directly utilize global confidence \cite{BDL}, and the others takes into account the imbalance in the distribution of different classes, using class-wise confidence or class-wise entropy \cite{prototypical_contrast}. 
The global-based methods select the top \(\zeta\)\% with the highest confidence or the lowest entropy from the predictions of all pixels as reliable pixels. In contrast, the class-wise-based methods initially segregate pixels belonging to different classes. Subsequently, for pixels of each class, it selects the top \(\zeta\)\% with the highest confidence or the lowest entropy as reliable pixels. To provide a more intuitive illustration of the distinctions between the global-based and class-wise-based methods, we conduct visualizations for global confidence and class-wise confidence, as presented in Fig. \ref{thresholding_method}. We perform experiments with all of these methods, as shown in Table \ref{tab: thresholding method}. To fairly compare different methods, we conduct multiple experiments with distinct values of \(\zeta\%\) and present the best results for each method in the Table \ref{tab: thresholding method}. We can find that using entropy yields the best results while utilizing class-wise confidence performs the poorest. Although class-wise-based methods are often superior to global-based methods in most research papers \cite{CBST, DSP}, they introduce more unreliable pseudo-labels due to the elevated error probability associated with predicting minority classes. The noise is magnified by contrastive learning, leading to a decrease in performance. Furthermore, numerous methods \cite{ESL, pixelIntraDA} have already indicated that using entropy tends to produce better results compared to confidence, and our experiments align with this viewpoint.

\noindent\textbf{Effectiveness of the target domain loss weight.} We train several models to evaluate the weight $\lambda_t$ in Eq (\ref{overall loss}), and the result is in Table \ref{tab: target domain loss weight}.
According to our experiments, the model performs best when $\lambda_t = 1$. This is because the weight of the target domain loss is related to 
$\kappa$, which is an adaptive value. If $\lambda_t$ is changed, it will significantly reduce the model`s performance.

\noindent\textbf{Effectiveness of the contrastive loss weight.} We evaluate how the weight $\lambda_c$ in Eq. \eqref{overall loss} affects the performance of GTA5 to Cityscapes adaptation. Table \ref{tab: table4} reveals the sensitivity of the results to diverse $\lambda_c$ values. Nonetheless, it is evident that results achieved with varying weights all surpass the baseline performance. We attribute this sensitivity to the imbalance between the contrastive and segmentation loss. The contrastive loss holds a larger magnitude compared to the segmentation loss in our experiments, thus the weight of the contrastive loss significantly influences the overall losses. Through our experiments,  we recommend using $\lambda_c = 0.1$ as a default setting.

\begin{table}[htbp]
	\doublespacing\caption{Analysis of the weighting factor $\lambda_t$ of target domain loss on GTA5 to Cityscapes. \label{tab: target domain loss weight}}
	\centering\small
	\begin{tabular}{l|cccccc}
		\toprule
		$\lambda_t$& 0.5& 0.8& 1.0& 1.2& 1.5& \\
		\midrule
		mIoU& 61.25& 61.19& \textbf{71.8}& 67.49&  67.69&  \\
		\bottomrule
	\end{tabular}
\end{table}

\begin{table}[htbp]
	\doublespacing\caption{Analysis of the weighting factor $\lambda_c$ of contrastive learning loss on GTA5 to Cityscapes. Note that $\lambda_c = 0$ means that the confidence loss is not applied. \label{tab: table4}}
	\centering\small
	\begin{tabular}{l|cccccc}
		\toprule
		$\lambda_c$& 0& 0.05& 0.1& 0.2& 0.3& 0.4\\
		\midrule
		mIoU& 68.3& 70.4& \textbf{71.8}&  71.0&  70.5& 70.2\\
		\bottomrule
	\end{tabular}
\end{table}

\noindent\textbf{Effectiveness of the number of anchor pixels and negative samples.} Through the modification of the number of anchor pixels and negative samples in our framework,  we observe varying results as depicted in Table \ref{tab: table5}. We find that an increased number of negative samples correlates with improved performance. This is because more negative samples approximate the true distribution more closely. Due to the limitation of computing resources, we take at most 512 negative samples.

\begin{table}[htbp]
	\doublespacing\caption{Analysis of the negative numbers on GTA5 to Cityscapes. \label{tab: table5}}
	\centering\small
	\begin{tabular}{l|ccccc}
		\toprule
		Anchor Numbers& 256& 256& 256&  128& 512\\
		Negative Numbers& 128& 256& 512& 512& 512\\
		\midrule
		mIoU& 69.9& 70.0&  \textbf{71.8}& 70.0& 70.3\\
		\bottomrule
	\end{tabular}
\end{table}

\noindent\textbf{Effectiveness of the proportion of reliable pixels.} We examine how the proportion of reliable pixels $\eta$ affects the performance in Table \ref{tab: table6}. We observe that $\eta = 0.5$ achieves the best result. If $\eta$ is too large, some incorrect samples may be generated. If $\eta$ is too small, some high-confidence samples may be ignored.
\begin{table}[htbp]
	\doublespacing\caption{Analysis of the proportion of reliable pixels on GTA5 to Cityscapes. \label{tab: table6}}
	\centering\small
	\begin{tabular}{l|ccccc}
		\toprule
		$\eta$& 0& 0.4& 0.5& 0.6& 0.7 \\
		\midrule
		mIoU& 68.3& 70.2& \textbf{71.8}& 70.5& 69.8\\
		\bottomrule
	\end{tabular}
\end{table}

\noindent\textbf{Effectiveness of the values of $r$.} We tested the performance of the model using Eq. \eqref{Target Negative Samples} with various values of $r$, and the results are presented in Table \ref{tab: table7}. By ranking the probabilities of all categories associated with a particular pixel, if category $c$ falls within the $r$ categories with the lowest probabilities, that pixel is considered as part of the negative sample set. Thus, the value of $r$ significantly impacts the robustness of the model. If the value of 
$r$ is relatively small, it may introduce noise. Conversely, if $r$ is relatively large, the distance between the pixels in the negative sample set and the target pixel in the feature space will be too far, making it ineffective for training. Based on our experiments, we found that $r = 4$ enables the model to achieve optimal robustness.

\begin{table}[htbp]
	\doublespacing\caption{Analysis of the values of $r$. \label{tab: table7}}
	\centering\small
	\begin{tabular}{l|ccccc}
		\toprule
		$r$& 3& 4& 5& 6 \\
		\midrule
		mIoU& 69.7& \textbf{71.8}& 69.6&  68.7\\
		\bottomrule
	\end{tabular}
\end{table}

\section{Conclusion}
In our study, we propose a novel contrastive learning approach to improve the effectiveness of domain adaptive semantic segmentation. First, we analyze the merits and demerits of existing contrastive learning methods and make improvements to address their limitations. Then, we explore the use of all target pixels to boost model performance, even though a substantial portion of pseudo-labels for target pixels is unreliable. Our method achieved remarkable results on two challenging benchmarks, surpassing previous methods by a significant margin. Extensive ablation studies are conducted to confirm the effectiveness of our approach. Moreover, we illustrate the adaptability of our approach to diverse architectures.

In the future, there are several approaches that can be employed to further enhance our method. First, our method requires a significant amount of computational resources during training, so reducing resource consumption can be considered a viable direction for improvement. Second, our method is still susceptible to errors caused by erroneous pseudo-labels, leading to error accumulation. Therefore, efforts may be directed toward denoising pseudo-labels.

\section{Data availability}
    The datasets analyzed during the current study are available at GTA5 dataset [\href{https://download.visinf.tu-darmstadt.de/data/from_games}{https://download.visinf.tu-darmstadt.de/data/from\_games}], Cityscapes dataset [\href{https://www.cityscapes-dataset.com}{https://www.cityscapes-dataset.com}], SYNTHIA dataset [\href{http://adas.cvc.uab.es/synthia}{http://adas.cvc.uab.es/synthia}], DarkZurich dataset [\href{https://www.trace.ethz.ch}{https://www.trace.ethz.ch}], and ACDC dataset [\href{https://acdc.vision.ee.ethz.ch}{https://acdc.vision.ee.ethz.ch}].

\section*{ACKNOWLEDGMENT}
This work was partially supported by the National Key Research and Development Program of China (2021YFB3301303), the National Natural Science Foundation of China (12105105, 62273149), the startup fund from East China University of Science and Technology under Grant JKH01241603 and the Programme of Introducing Talents of Discipline to Universities (the 111 Project) under Grant B17017.





\bibliographystyle{elsarticle-num} 
\bibliography{UnsupervisedDomainAdaptation}

\end{sloppypar}
\end{document}